# ParsEL 1.0: Unsupervised Entity Linking in Persian Social Media Texts

**Majid Asgari-Bidhendi, Farzane Fakhrian and Behrouz Minaei-Bidgoli**
Computer Engineering, Iran University of Science and Technology, Tehran, Iran
`majid.asgari@live.com, farzane_fakhrian@comp.iust.ac.ir, b_minaei@iust.ac.ir`

**Abstract**

In recent years, social media data has exponentially increased, which can be enumerated as one of the largest data repositories in the world. A large portion of this social media data is natural language text. However, the natural language is highly ambiguous due to exposure to the frequent occurrences of entities, which have polysemous words or phrases. Entity linking is the task of linking the entity mentions in the text to their corresponding entities in a knowledge base. Recently, FarsBase, a Persian knowledge graph, has been introduced containing almost half a million entities. In this paper, we propose an unsupervised Persian Entity Linking system, the first entity linking system specially focused on the Persian language, which utilizes context-dependent and context-independent features. For this purpose, we also publish the first entity linking corpus of the Persian language containing 67,595 words that have been crawled from social media texts of some popular channels in the Telegram messenger. The output of the proposed method is 86.94% f-score for the Persian language, which is comparable with the similar state-of-the-art methods in the English language.

## 1 Introduction

Entity linking (EL) is the task of linking a set of entities mentioned in a text to a knowledge base (KB). Entity linking is a developing field in natural language processing and plays an essential role in text analysis, information extraction, question answering, and recommender systems (Yan and Khurad, 2017). It also allows users to know about the background knowledge of entities in the text (Han et al., 2011). However, there are two types of ambiguities which make this task challenging.

Firstly, entities may have different names, even in a single document. For example, the name of a person can appear in the text as the first name, last name, or nickname. EL should link all of these names to a single entity in the knowledge base. Secondly, different entities may have the same name, but the entity linking system should be able to refer them to various entities from the knowledge base. Therefore, information about entities is crucial in choosing the correct entities (Han et al., 2011; Shen et al., 2015; Ganea et al., 2016).

Figure 1 presents a sample of disambiguation. For the word "apple", the mention of Apple entity can refer to multiple entities; however, only one of them refers to the correct entity. In almost all cases, based on the information provided in the context, only one of the candidate entities can be correct.

A knowledgebase is one of the fundamental components in entity linking systems. Generally, the knowledge base consists of a set of entities, information, semantic categories, and the relationship between entities. Knowledge bases used in EL systems should have some features such as public availability, machine readability, persistent identifiers, and credibility (Taufer and Straka, 2017). There are currently several knowledge bases for EL systems such as DBpedia (Auer et al., 2007), YAGO [1] (Suchanek et al., 2007), Freebase (Bollacker et al., 2008), and Probase (Wu et al., 2012). This study employs FarsBase (Asgari et al., 2019), which is the first multi-source knowledge base specially designed for the Persian language and includes more than 500,000 entities with 25 million relations between them. FarsBase can provide various information such as locations, persons, and organizations.

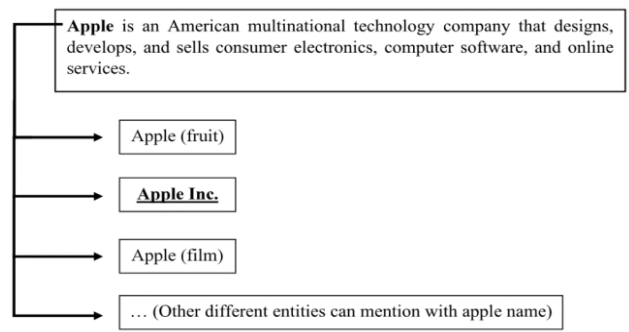

Figure 1: Entity disambiguation for entity mention apple in a text. The correct entity is underlined.

The rest of this paper is organized as follows. Section 2 discusses primary studies about entity linking. Section 3 introduces the new EL dataset for the Persian language. Section 4 describes the proposed approach for entity linking in the Persian language. Experimental results and the comparison of obtained results with the baseline method are discussed in section 5. The last section concludes this research and expresses our future work.

## 2 Related Works

In most cases, the entity linking process includes four subtasks, which are consistent with most of entity linking systems:

**Entity Recognition:** Most studies (Pershina et al., 2015; Ran et al., 2018; Hoffart et al., 2011) in entity linking employed existing algorithms for entity recognition, which had been provided by the other researches, and focused on the other three modules.

**Candidate Entity Generation:** This module proposes a set of candidate entities for every mentioned entity in the text (Shen et al., 2015; Wu et al., 2018). In this regard, most studies (Han et al., 2011; Guo et al., 2013; Gattani et al., 2013; Chong et al., 2017) used features such as redirect pages, disambiguation pages and hyperlinks in Wikipedia or mean relations in YAGO, to make a Name dictionary for each entity mention to map the entity mention to a set of candidate entities. Moreover, it becomes possible to get the candidates set from this Name dictionary (Shen et al., 2015).

**Candidate Entity Ranking:** In most cases, candidate entities are more than one. Therefore, candidate entities should be ranked by the EL system to find the proper entity from the knowledge base (Taufer and Straka, 2017). The EL system can apply two types of features for ranking candidates entities, which are Context-Independent Features and Context-Dependent Features (Shen et al., 2015). In the literature, the term "phase entity disambiguation" (Cucerzan, 2007; Dredze et al., 2010; Yamada et al., 2016) has the same meaning as candidate entity ranking. Additionally, both supervised and unsupervised methods can be used to achieve the results. Supervised ranking methods depend on the annotated training dataset, where its data annotation should be done manually.

**Unlinkable Mention Prediction:** In cases where entity mentions do not have any relevant entities in the knowledge base, unlinkable entity mentions can be separated from other entities and tagged as NIL. Different ways have been suggested by researchers to separate unlinkable mentions, namely ignoring unlinkable entity mentions (Han et al., 2011; Cucerzan, 2007; Han and Sun, 2012) NIL threshold (Yamada et al., 2016; Shen et al., 2012) and supervised machine learning techniques (Shen et al., 2015; Taufer and Straka, 2017; Zhang et al., 2010; Zhang et al., 2011).

### 2.1 Unsupervised Entity Linking

In some studies (Hoffart et al., 2011; Cucerzan, 2007), researchers release their manually annotated dataset for EL. These datasets are excellent benchmarks for the entity linking task. However, in the social media domain, making such a dataset is very hard, time-consuming, and costly. Also, most of the studies in EL focuses on the English language. Because of such a shortage, we decide to work on unsupervised methods.

Some researchers (Cucerzan, 2007; Han and Zhao, 2009; Chen et al., 2010; Xu et al., 2018) used Vector Space Model (VSM) (Salton et al., 1975) based methods for unsupervised candidate ranking. In this method, the first step is the calculation of the similarity between the vector representations of the entity mention and the candidate entity. The system links the candidate entity with the highest similarity to the entity mention. Their methods are different in the calculation of vector similarity and vector representation (Shen et al., 2015). Above that,

---

[1] YAGO (Yet Another Great Ontology)

| Research | Precision | Dataset |
|---|---|---|
| Chen et al.[26] | 71.2% | TAC-KBP2010 |
| Han and Zhao[25] | 76.7% | TAC-KBP2009 |
| Xu et al.[27] | 82% | Online Chinese Medical Text |
| Zhang et al.[29] | 91.2% | CoNLL |
| Cucerzan[10] | 91.4% | News dataset |
| Pan et al.[31] | 92.12% | News and discussion forum posts dataset |

Table 1: Results of main researches that used unsupervised learning for ranking.

because of colloquial text and misspelling problems, working on social media makes work harder.

Cucerzan (2007) used entity references mention in context and candidate entities articles to build vectors. To this end, the system will choose a candidate that maximizes vector similarity and have the same category as an entity mention. This system got 91.4% accuracy on a news dataset.

Chen et al. (2010) built the entity mention and candidate entities vectors based on the Bag of Words model by using the context of their article to capture word co-occurrence information and compute the similarity between them by TF-IDF similarity. They reported 71.2% accuracy on the TAC-KBP2010 dataset.

Han and Zhao (2009) used two types of similarity measures: the Wikipedia Semantic Knowledge-Based Similarity alongside Bag of Words based similarity. For generating vectors in the first similarity, the method detects Wikipedia concepts in candidate entities and context of the mentioned entity and then computes vector similarity of the entity mention and candidate entities using a weighted average of semantic relations between articles of Wikipedia concepts and the context of the mentioned entity. After that, these two types of similarity are merged, and the final similarity vector of the candidate entities is reported, and finally, the entity that maximizes this merged similarity is chosen. Their system achieves 76.7% accuracy on the TAC-KBP2009 dataset.

Xu et al. (2018) applied a linking approach for medical texts and exploit name similarity, entity popularity, category consistency, context similarity, and the semantic correlation between the entity mention and candidate entities, and rank candidate entities by combining these features. They call their ranking measure, Confidence Score. On average, their Confidence Score gets about 82% precision on their medical dataset.

Zhang et al. (2017) proposed an unsupervised bilingual entity linker inspired by Han and Sun (2011) and Yamada et al. (2016) researches. As we discussed before, they utilized a pre-built dictionary for the candidate generation, and after that, they used probabilistic generative methods to disambiguate the entities. Their system achieves 91.2% precision on the CoNLL dataset.

Pan et al. (2015) used Abstract Meaning Representation (AMR) (Banarescu et al., 2013) to select high-quality sets of entities for their similarity measure. They claimed that their representation using AMR could capture some contextual properties which are very critical and helpful for entities disambiguation without using training data. Next, for comparing the context of the entities, they used an unsupervised graph to get final results and reported 92.12% precision on a dataset annotated from news and discussion forum posts.

Table 1 summarizes the final results of all the above studies. Researchers have studied entity linking in the Persian language less than the English language. To our knowledge, the proposed solution is the first Persian entity linker.

## 3 Dataset

In this research, we introduce the ParsEL-Social corpus, which is constructed from social media contents derived from 10 Telegram channels in 10 different categories: sport, economics, gaming, general news, IT news, travel, art, academic, entertainment, and health.

| Dataset | Count |
|---|---|
| Documents | 4,263 |
| Sentences | 6,160 |
| Words | 67,595 |
| Entities | 19,831 |
| Candidates | 145,148 |
| Words per article | 15.9 |
| Entities per article | 4.7 |
| Candidates per Entity mentions | 7.3 |

Table 2: ParsEL-Social Dataset properties.

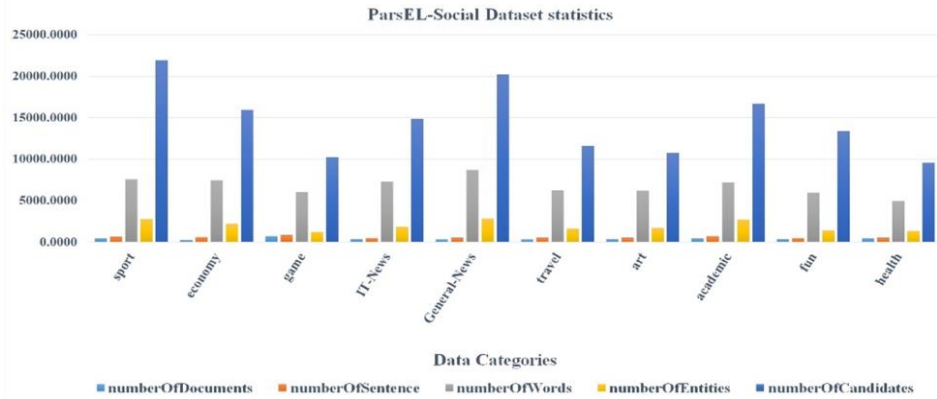

Figure 2: ParsEL-Social Dataset statistics per category

Table 2 summarizes the statistics of the ParsEL-Social dataset, such as the number of posts, words and entities, the average number of words and entities in each post, and the average number of candidates for each entity.

Moreover, Figure 2 shows the statistics of the ParsEL Social dataset for each category. Should be noted, the numbers of documents in the sport and academic categories are higher than in other categories because the posts are shorter in the datasets. The number of sentences is fewer in the game, travel, and health categories because longer sentences are used in the posts. The corpus distributes an equal number of words for all the categories. Texts in the sport, general news, and academic categories have a higher number of entities, but the differences are not remarkable as the entities are not restricted to named entities. Finally, the number of candidates is higher in the sport, general news, and academic categories; therefore, these types of texts have more ambiguous words.

## 4 Proposed Entity Linking Method

Like other entity linking methods, the proposed method focuses on the candidate generation, the ranking, and unlinkable mention predictions.

Initially, the proposed method uses FarsBase for the candidate entity generation. For each entity in FarsBase, a predicate named "variantLabel" obtains its values from Wikipedia redirect pages and has different versions of the name of entities. For every word, the algorithm extracts all possible entities based on its variantLabel in the FarsBase. By using this method, we can generate a candidate set for each entity mention in the candidate ranking of the next step.

In the candidate ranking phase, the goal is to link each entity mention to only one knowledge base entity from the candidate set. We utilize both of the context-dependent and context-independent features in the ranking step. Context-dependent features rely on the context where entity mention appears, but context-independent features are independent of context and rely on entity mention and candidate entities (Shen et al., 2015).

First of all, we use these following heuristics to remove some of the inappropriate candidates:

**Type Checking:** The system checks the types of entities and eliminates candidates whose type is not the same as the entity mention of the candidate set.

**POS Tags:** Following the type checking, a built-in POS tagger from our knowledge base is used to tag sentences surrounding the entity mentions and eliminate the entities that have POS tag different from the entity mentions.

**The popularity of entities:** Entities with the same mention have different popularity (Shen et al., 2015). Take Tehran as an example; Tehran (city) is much more used than Tehran University. Therefore, in these cases, rare entities are ignored using a manually created list.

**Class-specific Filters:** Some entities have a very generic name that may cause a high level of ambiguity. For instance, "چهل سالگی" ("At the age of 40") is an Iranian movie while it can be as a part of a general sentence, e.g., "Vahid died at the age of 40". Such names are widespread in artworks (e.g., movies or books) and a limited number of the other specialized classes. To improve the disambiguation process, we will look for more evidence in the context using a reference list if the candidate entity belongs to individual classes. Considering the above example, "At the age of 40", the surrounding context containing phrases such as channel, cinema, ticket, and a movie is required. Otherwise, the algorithm multiplies the real rate of the candidate by a predefined constant number between 0 and 1 based on each case.

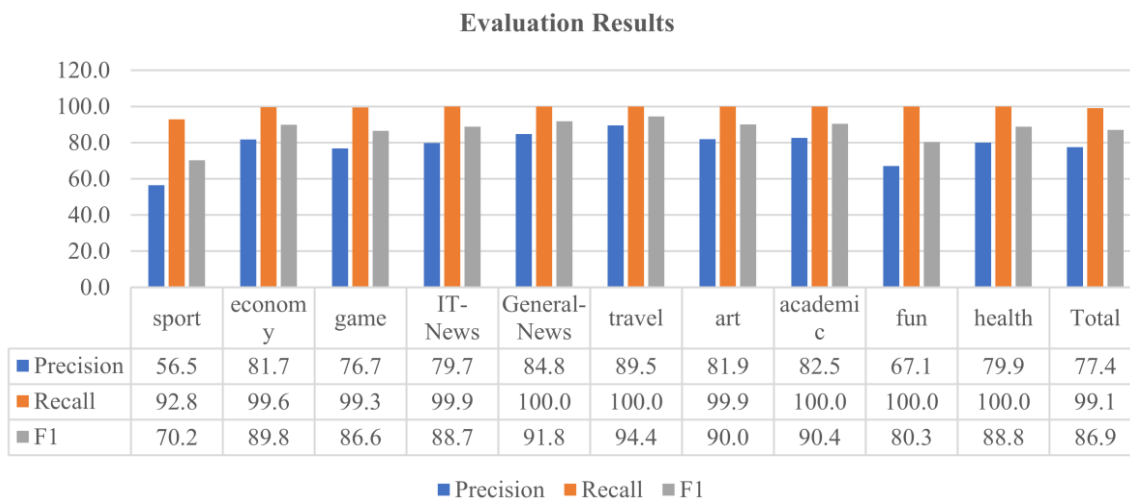

Figure 3: Entity linking results using the proposed method on ParsEL-Social dataset.

After removing some of the incorrect candidates, the system scores the remaining candidates. The scoring method employs context-dependent features and follows the four following steps:

- **Context Score:** The first step is to compute the cosine similarity between the words of the context of the entity mention and the textual context of the corresponding Wikipedia article of candidate entities. This step ignores the stop words in the Persian language.
- **Graph Score:** In the next step, candidates are scored based on the number of hyperlinks between all candidate entities in their corresponding Wikipedia articles.
- To rank the candidates, we merge the context score and graph score.
- Finally, the system links the candidate entity with the highest score to the entity mention. Other entities will be added to the entity mention's "ambiguity-list" to persist the rejected candidates for possible future applications such as error checking.

After candidate generation and ranking, the NIL threshold method (Shen et al., 2015; Yamada et al., 2016; Shen et al., 2012) is used for unlinkable mention prediction. In this method, if the score of the top-ranked candidate entity is lower than the pre-defined threshold, the entity mention will be tagged as NIL, and the system will add all of the candidate entities to the ambiguity-list.

## 5 Results and Evaluation

We evaluate the proposed unsupervised method (ParsEL 1.0[1]) on the ParsEL-Social dataset, and the results are reported in Figure 3 for each category using precision, recall, and F1 measures. The proposed method is comparable with the state-of-the-art unsupervised methods on TAC-KBP datasets, and the results are acceptable for the first Persian entity linker.

Table 3 compares ParsEL with a baseline method. For the baseline, we use Babelfy[2] entity linking, which works based on BabelNet 3.0[3]. In the first step. We run Babelfy on our dataset by public APIs of Babelfy. Babelfy returns all of the BabelNet synsets for each token in the text. Each synset is linked to some sources such as Wordnet or Wikipedia articles in different languages. Synset sources are available on the page of the synset or public BabelNet APIs. Each Wikipedia article in the Persian language corresponds to a FarsBase entity. Since BabelNet merges multiple sources to construct its synsets and, on the other hand, FarsBase is based on Persian Wikipedia, we only get Persian Wikipedia sources for each synset and convert it to FarsBase links. Therefore, each BabelNet synset can be linked to its corresponding entity in the FarsBase knowledge graph. As it was discussed earlier, BabelNet synsets are not extracted only from Persian Wikipedia, thus, comparing the reported recall rate with the ParsEL is not wholly impartial, and it is normal for baseline recall to be lower.

---

[1] ParsEL is the entity linker Raw-Text Extractor Module of the FarsBase project. FarsBase is an open-source system and is available in https://github.com/IUST-DMLab/farsbase-kg.

[2] http://babelfy.org
[3] https://babelnet.org

| Category | Baseline P | Baseline R | Baseline F1 | ParsEL P | ParsEL R | ParsEL F1 |
|---|---|---|---|---|---|---|
| Sports | 0.5111 | 0.4720 | 0.4908 | 0.5647 | 0.9282 | 0.7022 |
| Economy | 0.4855 | 0.5676 | 0.5234 | 0.8174 | 0.9961 | 0.8979 |
| Game | 0.3790 | 0.5430 | 0.4464 | 0.7669 | 0.9934 | 0.8656 |
| IT News | 0.4061 | 0.4375 | 0.4212 | 0.7974 | 0.9994 | 0.8870 |
| General News | 0.4638 | 0.5080 | 0.4849 | 0.8476 | 1.0000 | 0.9175 |
| Travel | 0.4572 | 0.2297 | 0.3058 | 0.8946 | 1.0000 | 0.9444 |
| Art | 0.4576 | 0.2746 | 0.3433 | 0.8193 | 0.9987 | 0.9002 |
| Academic | 0.5757 | 0.5296 | 0.5517 | 0.8252 | 1.0000 | 0.9042 |
| Fun | 0.4279 | 0.4531 | 0.4402 | 0.6707 | 1.0000 | 0.8029 |
| Health | 0.4830 | 0.4818 | 0.4824 | 0.7987 | 1.0000 | 0.8881 |
| Total | 0.4716 | 0.4546 | 0.4630 | 0.7744 | 0.9911 | 0.8694 |

Table 3: Comparing ParsEL with the baseline algorithm.

## 6 Conclusions and Future Work

In this paper, we presented ParsEL, an entity linker for the Persian language, which uses the FarsBase knowledge graph as its dataset. The results show that the precision of ParsEL is comparable with the entity linkers in other languages. Using multiple heuristics enables ParsEL to compete with state-of-the-art unsupervised methods for entity linking even in other languages.

In future work, we plan to annotate a larger dataset for supervised approaches. Deep learning has improved entity linking results in recent years, which can be the correct choice for the next versions of ParsEL. Besides, extracted links from a piece of text must have reasonable relationships. A post-processing phase can investigate these relationships and improve the overall results. Using entity or word embedding also can improve the proposed method for entity linking in the Persian language.